\documentclass[conference]{IEEEtran}
\IEEEoverridecommandlockouts
\usepackage{cite}
\usepackage{amsmath,amssymb,amsfonts}
\usepackage{algorithmic}
\usepackage{graphicx}
\usepackage{multicol}
\usepackage{multirow}
\usepackage{textcomp}
\usepackage{xcolor}
\usepackage{float}
\usepackage{booktabs}
\usepackage{marvosym}
\usepackage{stfloats}
\usepackage{bbm}
\usepackage{siunitx}
\usepackage{svg}
\usepackage{caption}
\usepackage{float}
\def\BibTeX{{\rm B\kern-.05em{\sc i\kern-.025em b}\kern-.08em
    T\kern-.1667em\lower.7ex\hbox{E}\kern-.125emX}}

\makeatletter

\def\ps@IEEEtitlepagestyle{%
  \def\@oddfoot{\mycopyrightnotice}%
  \def\@evenfoot{}%
}
\def\mycopyrightnotice{%
  {\footnotesize 979-8-3503-3748-8/23/\$31.00~\copyright~2023 IEEE\hfill} 
  \gdef\mycopyrightnotice{}
}

\begin{document}

\title{Auto-Augmentation Contrastive Learning for Wearable-based Human Activity Recognition}

\author{
\IEEEauthorblockN{Qingyu Wu\textsuperscript{1,2,3}, Jianfei Shen\textsuperscript{1,2,3,4}, Feiyi Fan\textsuperscript{1,2,3}, Yang Gu\textsuperscript{1,2,3,5}, Chenyang Xu\textsuperscript{6}, Yiqiang Chen\textsuperscript{1,2,3,4,5,\Letter}}
\IEEEauthorblockA{$^1$\textit{Institute of Computing Technology, Chinese Academy of Sciences, Beijing, China}\\
$^2$\textit{University of Chinese Academy of Sciences, Beijing, China}\\
$^3$\textit{The Beijing Key Laboratory of Mobile Computing and Pervasive Device, Beijing, China}\\
$^4$\textit{Shandong Academy of Intelligent Computing Technology, ShanDong, China}\\
$^5$\textit{Peng Cheng Laboratory, Shenzhen, China}\\
$^6$\textit{Institute of Robotics and Autonomous Systems, Tianjin University, Tianjin, China}\\
\{wuqingyu21s, shenjianfei, fanfeiyi, guyang, yqchen\}@ict.ac.cn, chyond.xu@gmail.com}
}

\maketitle



\begin{abstract}
For low-semantic sensor signals from human activity recognition (HAR), contrastive learning (CL) is essential to implement novel applications or generic models without manual annotation, which is a high-performance self-supervised learning (SSL) method. However, CL relies heavily on data augmentation for pairwise comparisons. Especially for low semantic data in the HAR area, conducting good performance augmentation strategies in pretext tasks still rely on manual attempts lacking generalizability and flexibility. To reduce the augmentation burden, we propose an end-to-end auto-augmentation contrastive learning (AutoCL) method for wearable-based HAR. AutoCL is based on a Siamese network architecture that shares the parameters of the backbone and with a generator embedded to learn auto-augmentation. AutoCL trains the generator based on the representation in the latent space to overcome the disturbances caused by noise and redundant information in raw sensor data. The architecture empirical study indicates the effectiveness of this design. Furthermore, we propose a stop-gradient design and correlation reduction strategy in AutoCL to enhance encoder representation learning. Extensive experiments based on four wide-used HAR datasets demonstrate that the proposed AutoCL method significantly improves recognition accuracy compared with other SOTA methods.
\end{abstract}







\section{Introduction}
With the development of mobile and wearable computing, human activity recognition (HAR) using the inertial measurement unit (IMU) on wearable devices is researched in diverse areas. And the applications and services of wearable-based HAR are becoming ubiquitous, such as assisting life, context-aware, anomaly detection, human-computer interaction, etc. These researches use sensors placed at different body locations to perform data mining and inference, which is critical to human-centered care and enables ubiquitous intelligence. Wearable-based HAR has traditionally been explored through sensor-based methods leveraging signal processing algorithms or conventional machine learning, lacking robustness, especially for complex activities. The landscape of HAR research has been significantly reshaped in recent years by introducing deep learning models such as convolutional neural networks, recurrent neural networks, attention-based networks, etc., effectively elevating precision and robustness. These methods are mostly based on supervised learning, which requires recollecting data and retraining across different tasks because the activities, sensor types, sensor positions, etc., are varied, which means a heavy manual annotation burden.



\begin{figure}[t]
    \centering
    \includegraphics[scale=0.335]{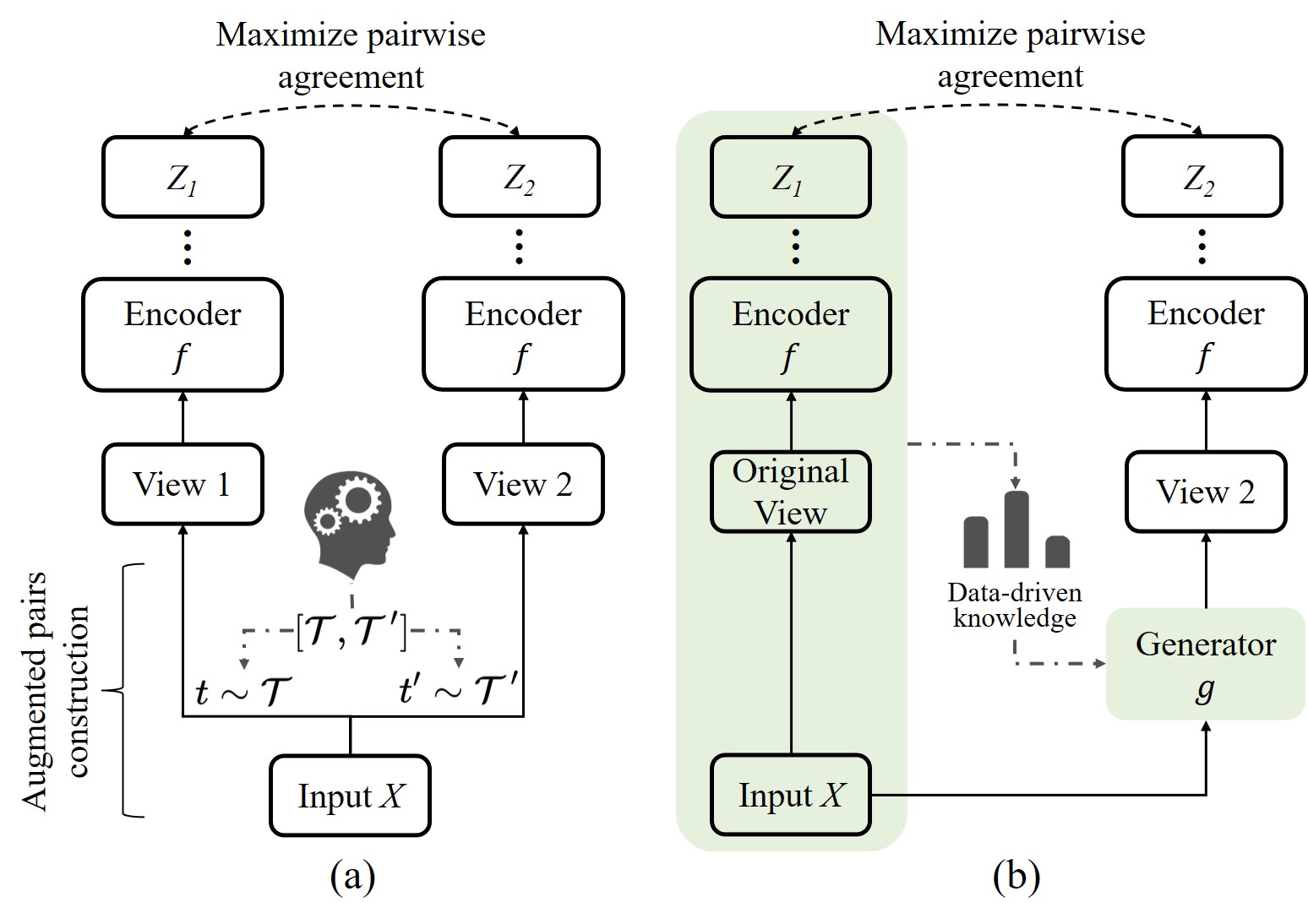}
    \vspace{-0.7cm}
    \caption{(a) The contrastive learning framework relies on data augmentation strategies to construct different views of each sample to maximize pairwise agreement \cite{SimCLR, reviewCL}. (b) The illustration of AutoCL's exploring of data-driven augmentation contrastive learning method.}
    \label{fig:introduction}
    \vspace{-0.3cm}
\end{figure}

Therefore, self-supervised learning (SSL) without manual annotation, especially contrastive learning (CL) becomes essential, which is a high-performance SSL method. It constructs pairwise samples of the original sample in the pretext task and learns the latent space by decreasing the positive sample distance and increasing the negative sample distance through CL loss functions. At the early stage of CL development, the need to construct and accumulate negative samples to enlarge the receptive field of the model led to approaches with the memory bank design \cite{InstDisc}, momentum encoder design \cite{moco}, etc., which caused much extra computational and storage overhead. Starting with SimCLR's demonstration of the option to directly use samples in the mini-batch as negative samples while training and conducting excellent performance as well \cite{SimCLR}, the construction of negative pairs was gradually simplified. 

However, the construction of positive samples still requires. Current mainstream contrastive learning approaches, as in Figure \ref{fig:introduction}(a), require manual selection and implementation of different augmentation strategies to construct positive augmented views for maximizing pairwise agreement. Especially in wearable HAR, data augmentation strategies are not as intuitive as they are for highly semantic data such as images, so it is challenging to intuitively select or construct efficient data augmentation approaches for specific HAR tasks. Previous studies have shown that the optimal augmentation strategy is inconsistent across different tasks due to the varying number of sensors, sampling rate, modality, etc. \cite{CL-HAR}. This makes CL in wearable HAR for data augmentation of low semantic data from sensors still manually dependent and lacks flexibility and generalizability.



In this paper, we explore an auto-augmentation contrastive learning method in HAR, which achieves end-to-end self-supervised contrastive learning of HAR sensor signals without additional manual construction of augmentation strategies by the idea of generating augmented samples through the knowledge of the original sample itself, as illustrated in Figure \ref{fig:introduction}(b). Auto-augmentation is performed directly during the model training process while achieving CL. Specifically, AutoCL employs a Siamese network architecture that shares the parameters of the backbone as \cite{NNCLR, SimCLR, SimSiam}. A generator is embedded in the architecture to learn auto-augmentation from the feature embedding of the backbone instead of the original data to reduce the effect of noise and redundant information in raw sensor data. Finally, the model implements contrastive learning based on the original and auto-augmented samples by maximizing pairwise agreement. And we propose stop-gradient and correlation reduction strategies in AutoCL to learn a better latent space representation. The contributions of our paper are summarized as follows:


\begin{itemize}

\item We explore an auto-augmentation contrastive learning method to eliminate the data augmentation burden of low semantic sensor signals in HAR. To the best of our knowledge, this is the first study that conducts end-to-end auto-augmentation CL for wearable HAR.

\item An AutoCL framework with a generator embedded is designed and uses the stop-gradient design and correlation reduction strategy to enhance encoder representation learning. It achieves state-of-the-art performance compared to the baselines with current HAR data augmentation strategies. The evaluation results further demonstrate the prospects of the data-driven AutoCL framework.

\item Our insightful discussion provides additional analyses of the effectiveness of our approach by architecture empirical study. Additionally, we visualize the auto-augmented samples, which shows the space of auto-augmentation operations is far more than the human-specified strategy, with the capability for better performance.

\end{itemize}


\section{Related Work\label{sec:RELATEDWORK}}
This section presents an overview of the existing literature on human activity recognition and contrastive learning techniques, encapsulating prior efforts in these areas.

\subsection{Human Activity Recognition}
Considerable research in recent years has targeted wearable human activity recognition, with a primary focus on extracting features through deep neural networks, such as convolutional networks, residual networks, attention-based networks, etc. \cite{deeplearningHAR0, deeplearningHAR1, deeplearningHAR2}. These models show commendable accuracy provided that many training samples are available. However, their performance drops in scenarios where labels are scarce. This problem has been approached through semi-supervised learning techniques, which employ unlabeled training samples that are more readily accessible \cite{semilearning0, semilearning1}. Furthermore, weakly-supervised learning strategies employ minimal inaccurate or coarse labels to alleviate this issue \cite{weaklearning, weaklearning1}. Transfer learning methods have also been proposed to relay helpful information from label-abundant source domains to label-sparse target domains \cite{translearning0}.

Recently, self-supervised learning, particularly contrastive learning, has shown exemplary capabilities in feature learning without supervision \cite{ssl0, COCOA}. Many recent investigations have been into various forms of data transformation in contrastive learning. \cite{augmentationmethods} examined several data augmentation techniques designed for wearable sensors, replacing image augmentation operators in the SimCLR model. \cite{stfcl} implemented both time-domain and frequency-domain augmentation techniques in SimCLR, adapting the encoder from \cite{yao2019stfnets} to accommodate features in time and frequency domains. The temporal and contextual contrasting method with jittering and permutation augmentation is proposed to incorporate the temporal characteristics of time series into contrastive learning \cite{TSTCC}.

\subsection{Contrastive Learning}
Driven by the constraints of manual annotation, contrastive learning generates alternate views of the same sample to form positive pairs in a label-free environment. Methods like SimCLR derive two batches of positive samples from the original samples and compare them against a pool of other samples \cite{SimCLR}. This strategy is rooted in the principle of sample discrimination, where each sample is considered a unique category. Besides generating positive samples from a single sample, some methods assign samples from the same cluster or nearest neighbors as positives \cite{clusterCL0}. Techniques such as BYOL \cite{BYOL} and SimSiam \cite{SimSiam} based on the dual-stream architecture have demonstrated competitive performance without requiring negative samples, thereby circumventing the necessity for large numbers of such samples. Recent works have provided a review of existing approaches \cite{reviewCL}.

Even with the considerable enhancements brought about by contrastive models, understanding the underlying factors, especially pairwise augmentation, is a nascent research area. This has instigated studies exploring the mechanisms that drive the success of contrastive learning \cite{studyCL0, studyCL1, studyCL2}. Existing reliance on contrast learning for pair construction introduces an additional task of non-intelligent, artificial sample augmentation in the pretext task, which needs more flexibility and generalizability. This issue is especially evident for low-semantic sensor signals, where creating semantically consistent sample pairs is a significant challenge \cite{augmentationfree}. In wearable HAR, some studies have adopted a multimodal approach to enable cross-channel sensor signals comparison without additional augmentation \cite{COCOA}. However, it is constrained by the limited number of sensors available in practical applications. To overcome the limitations of data augmentation in existing contrastive learning methods, our work aims to conduct an end-to-end contrastive learning study from an auto-augmentation standpoint for wearable HAR.

\begin{figure*}[tp]
  \centering
  \includegraphics[scale=0.35]{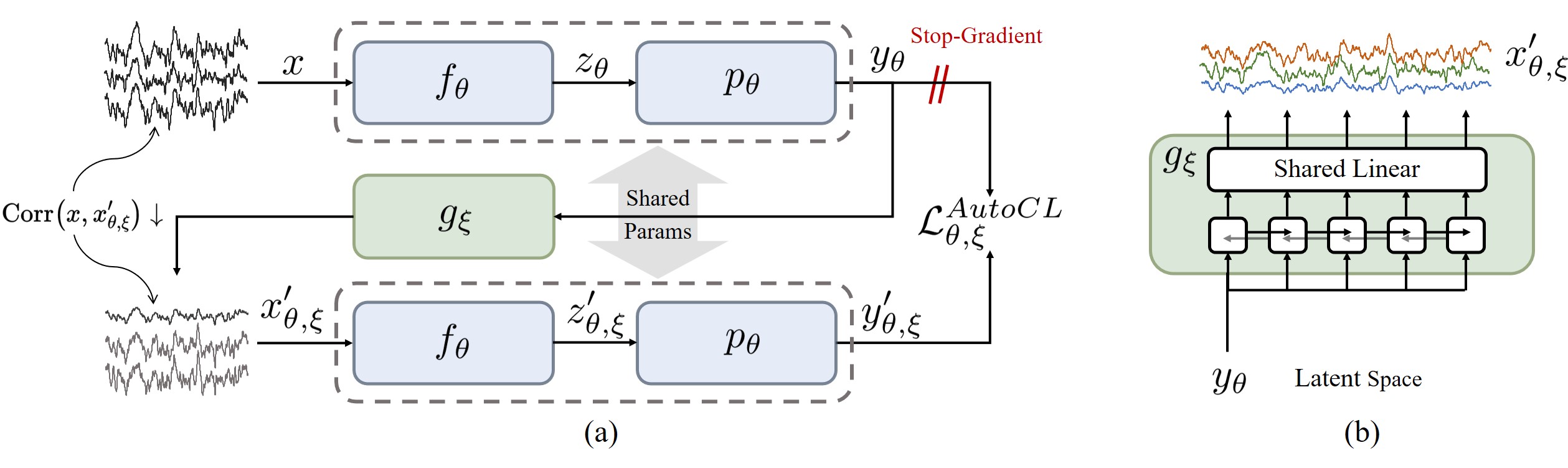}
  \vspace{-0.3cm}
  \caption{Overall of AutoCL Methodology. (a) AutoCL architecture. (b) Implementation of the generator.}
  \label{fig:methodology}
  \vspace{-0.3cm}
\end{figure*}

\section{Methodology}
In this section, we propose an Auto-Augmentation Contrastive Learning method for HAR, namely AutoCL. We are given the pretraining dataset $\mathcal{D}_{pret} = \{x_{i}^{pret} | i = 1, . . . , N \}$ of unlabeled human activity time series samples. Let $\mathcal{D}_{tune} = \{x_{i}^{tune} | i = 1, . . . , M \}$ be a fine-tuning dataset of labeled human activity time series samples. The goal is to use $\mathcal{D}_{pret}$ to pre-train an encoder $f$ so that by fine-tuning model parameters on $\mathcal{D}_{tune}$, the fine-tuned model produces good representations $f(x_i^{tune})$ for every $x_i^{tune}$.

\subsection{Motivation}\label{sec:Motivation}

Many successful contrastive learning approaches based on Siamese network architecture rely on selecting suitable augmentation strategies to generate augmented samples for comparison, aiming to maximize pairwise agreement. The AutoCL method, in its pursuit of data-driven automatic augmentation, is inspired by the Squeeze-and-Excitation (SE) attention mechanism \cite{SE} (self-guiding its own attention weighting). Intuitively, AutoCL aspires to leverage the knowledge from the original samples to drive a generator for producing augmented samples (self-guiding its own augmentation strategy) and maximize the pairwise agreement via the contrastive learning loss.

The most crucial consideration in this process is whether to use the original samples directly as the generator input or use the feature embeddings from the backbone network as the generator's input for auto-augmentation. Due to the low semantic information and susceptibility to noise interference inherent in sensor signals, extracting valuable features and generating augmented data is challenging. This problem is indicated in the HAR sensor signals generation study by Wang et al. \cite{SensoryGANs1, SensoryGANs}, who suggested that better performance can be achieved by generating HAR sensor signals based on the feature embeddings.

Therefore, we explore the architecture design of AutoCL: using the feature embeddings from the backbone network to guide the generator for auto-augmentation in self-supervised contrastive learning. Subsequently, original samples and auto-augmented samples constitute sample pairs to guide the backbone network for more effective contrastive learning. The details of AutoCL architecture are as follows. We further provide the empirical study in Section \ref{sec:EmpiricalStudy}, which attests to the rationality and advancement of the AutoCL architecture.

\subsection{Description of AutoCL}
AutoCL’s goal is to learn a representation $z_{\theta}$, which can then be used for downstream tasks. As described previously, AutoCL uses a Siamese network and a generator network to learn, as shown in Figure \ref{fig:methodology}. The Siamese network is defined by a set of weights $\theta$ and is comprised of two stages: an encoder $f_{\theta}$, a projector $p_{\theta}$. The generator $g_{\xi}$ uses a set of weights $\xi$ and is obtained by training together with the Siamese network's weights $\theta$.

Given a set of unlabeled time series samples $\mathcal{D}_{pret}$, a sample $x \sim \mathcal{D}_{pret}$ sampled uniformly from $\mathcal{D}_{pret}$. AutoCL produces augmented sample $x'_{\theta, \xi} \triangleq g_{\xi}(y_{\theta})$, where $y_{\theta}$ is from $x$ embedded by $f_{\theta}$ and $p_{\theta}$. For the first input of the Siamese network, the network outputs a representation $z_{\theta} \triangleq f_{\theta}(x)$ and a projection $y_{\theta} \triangleq p_{\theta}(z_{\theta})$. The other stream of the Siamese network outputs $z'_{\theta, \xi} \triangleq f_{\theta}(x'_{\theta, \xi})$ and the projection $y'_{\theta, \xi} \triangleq p_{\theta}(z'_{\theta, \xi})$ from the second input $x'_{\theta, \xi}$.
Finally, we define the normalized temperature-scaled cross-entropy (NT-Xent) loss function between the projections $y_{\theta}$ and $y'_{\theta, \xi}$ \cite{SimCLR}: 

\begin{equation}
\mathcal{L}_{\theta, \xi}=-\log{\frac{\exp(\text{sim}(y_{\theta, i},y'_{\theta, \xi, i})/\tau)}{\sum_{k=1}^{2N}{\mathbbm{1}}_{[k\neq i]}\exp(\text{sim}(y_{\theta, i},y_{\theta, k})/\tau)}} 
\end{equation}

where $N$ is the batch size, $\tau$ is a temperature hyper-parameter that controls the smoothness of the distribution, $\text{sim}(y_{\theta, i}, y'_{\theta, \xi, i})$ is the cosine similarity between the embeddings of the original and augmentation projection, and $\mathbbm{1}$ is the indicator function that equals $1$ if $k \not= i$ and $0$ otherwise. In our work, $\tau$ is $0.1$.

\subsection{Correlation Reduction}

During the training process of AutoCL, we believe that relying solely on generated samples for training $g_{\xi}$ towards the CL loss function may lead to overfitting and loss of the capacity to learn semantics from sensor signals better. If the results lead to $x'{\theta, \xi} \rightarrow x{\theta}$, it reduces the comparative difficulty for contrastive learning, approximating the training process of an auto-encoder, which has been shown to perform far less effectively than contrastive learning \cite{CLvsAE}. It is indicated that the smaller the correlation between positive samples, the better the performance of contrastive learning \cite{CL-HAR}. 
We added a term to the loss function, ${\frac{\sum_{i=1}^{N}(x_{i}-{\overline{{x}}})(x'_{\theta, \xi, i}-{\overline{{x'_{\theta, \xi}}}})}{{\sqrt{\sum_{i=1}^{N}(x_{i}-{\overline{{x}}})^{2}}}{{\sqrt{\sum_{i=1}^{N}(x'_{\theta, \xi, i}-{\overline{{x'_{\theta, \xi}}}})^{2}}}}}}$, attempting to maximize the diversity between pairs of positive samples during contrastive learning, thus enhancing the performance of contrastive learning. The final loss function of AutoCL $\mathcal{L}_{\theta, \xi}^{AutoCL}$ is as follows:

\begin{equation}
\begin{split}
\mathcal{L}_{\theta, \xi}^{AutoCL}=-\log{\frac{\exp(\text{sim}(y_{\theta, i},y'_{\theta, \xi, i})/\tau)}{\sum_{k=1}^{2N}{\mathbbm{1}}_{[k\neq i]}\exp(\text{sim}(y_{\theta, i},y_{\theta, k})/\tau)}} \\+ {\frac{\sum_{i=1}^{N}(x_{i}-{\overline{{x}}})(x'_{\theta, \xi, i}-{\overline{{x'_{\theta, \xi}}}})}{{\sqrt{\sum_{i=1}^{N}(x_{i}-{\overline{{x}}})^{2}}}{{\sqrt{\sum_{i=1}^{N}(x'_{\theta, \xi, i}-{\overline{{x'_{\theta, \xi}}}})^{2}}}}}} 
\end{split}
\end{equation}

\subsection{Stop-Gradient}

During the training process, in order to make the generation process of the generator $g_{\xi}$ depend on the feature embedding of $f_{\theta}$ and $p_{\theta}$, and let the generated sample $x'_{\theta, \xi}$ through $f_{\theta}$ and $p_{\theta}$ for CL to conduct recurrent training of $f_{\theta}$, we stop the gradient in $y_{\theta}$ as in Figure \ref{fig:methodology}(a), resulting in the final gradient calculation as follows:

\begin{equation}
\nabla \mathcal{L}_{\theta, \xi}^{A u t o C L}=\frac{\partial \mathcal{L}_{\theta, \xi}}{\partial p_\theta} \frac{\partial p_\theta}{\partial f_\theta} \frac{\partial f_\theta}{\partial g_{\xi}} \frac{\partial g_{\xi}}{\partial p_\theta} \frac{\partial p_\theta}{\partial f_\theta} \frac{\partial f_\theta}{\partial \theta, \xi}
\end{equation}

Finally, We perform a stochastic optimization step to minimize $\mathcal{L}_{\theta, \xi}^{AutoCL}$ with respect to $\theta$ and $\xi$. AutoCL’s dynamics are summarized as follows:

\begin{equation}
{\theta, \xi} \leftarrow\mathrm{optimizer}\bigl(\theta, \xi, \nabla{\mathcal{L}}_{\theta,\xi}^{\mathrm{AutoCL}}, lr) 
\end{equation}
where $\mathrm{optimizer}$ is an optimizer and $lr$ is a learning rate.

\subsection{Implementation Details}\label{sec:ImplementationDetails}
The encoder, projector, generator, and optimization implementation details are introduced in this section. Specifically, the inputs $x, x'_{\theta, \xi} \in \mathbb{R}^{N \times W \times C}$, which are sensor signals in HAR. $N$ represents the batch size. $W$ represents the time window. $C$ represents the number of signal channels.

\subsubsection{Encoder}\label{sec:encoder}
For the realization of the encoder $f_{\theta}$, we opt for a 3-layer fully convolutional network (FCN) \cite{FCN}. Each layer of the FCN consists of a 1-dimensional (1-D) convolutional layer, accompanied by batch normalization, ReLU activation, max pooling, and a 10\% dropout operation. Every layer utilizes a convolutional kernel of size 8 with 1 stride and a pooling kernel of size and stride both of 2.

\subsubsection{Projector}\label{sec:projector}
The projector $p_{\theta}$, an essential structure after the encoder in contrastive learning, assists in the training of the encoder throughout the self-supervised process, amplifying the robustness of the encoder \cite{SimCLR}. In our research, $p_{\theta}$ adopts a Multi-Layer Perceptron (MLP) design, including a fully connected layer incorporating 256 neurons, batch normalization, and ReLU activation function, and concluding with another fully connected layer hosting 128 neurons with a softmax activation function.

\subsubsection{Generator}\label{sec:generator}
The generator $g_{\xi}$, as depicted in Figure \ref{fig:methodology}(b), duplicates the feature embedding $y_{\theta}$ over time and introduces it into a 3-layer bidirectional gate recurrent unit (BiGRU). The hidden layer's size within this network amounts to $4 \times C$. A final global linear transformation is executed via a 1-D group convolutional layer with kernel size and stride of 1, mapping the $4 \times C$ channels to $C$ by $C$ group, thereby producing $y'_{\theta, \xi}$. Notably, batch normalization and ReLU activation are performed before feature input the generator and are used to avoid overfitting after the MLP projection module.

\subsubsection{Optimization}\label{sec:optimization}
The optimization process is carried out employing the Adam optimizer with a learning rate and weight decay both set at $\num{1e-3}$.

\section{Experiments}
In this section, we evaluate the performance of AutoCL on HAR datasets. To evaluate the performance of AutoCL for automatic augmentation through end-to-end self-supervised learning, we adopt widely adopted advanced data augmentation methods in HAR: scale, permutation, noise, and their combination using on state-of-the-art CL methods\cite{TSTCC, SimCLR, SimSiam, BYOL, NNCLR}. Moreover, we evaluate the reduced correlation strategy and the stop-gradient design for ablation experiments to assess its effectiveness. 

\begin{table}[htbp]
\centering
\footnotesize
\caption{Briefly description of experimental HAR datasets}
\label{tab:Dataset}
\setlength{\tabcolsep}{3pt}
\renewcommand{\arraystretch}{1.2}
\begin{tabular}{ccccc}\toprule
\multicolumn{1}{c}{Description}& 
\multicolumn{1}{c}{PAMAP2}&
\multicolumn{1}{c}{UCIHAR}&
\multicolumn{1}{c}{UTD-MHAD}&
\multicolumn{1}{c}{DSADS}
\\
\cmidrule(lr){1-5}
Number of Subjects &
$9$ & 
$30$ & 
$8$ & 
$8$\\
Number of Activities &
$12$ & 
$6$ & 
$27$ &
$19$\\
Number of Channels &
$36$ &
$9$ &
$6$ &
$45$\\
Sample Rate & 
$100$Hz & 
$50$Hz &
$50$Hz &
$25$Hz\\
\cmidrule(lr){1-5}
Window Size & 
$128$ & 
$128$ & 
$128$ &
$128$\\
Overlap & 
$50\%$ & 
$50\%$ & 
$50\%$ &
$50\%$\\
\bottomrule
\end{tabular}
\vspace{-0.3cm}
\end{table}

\subsection{Dataset Description}
We evaluate the effectiveness of our approach on four widely-used public datasets, summarised in Table \ref{tab:Dataset}. The PAMAP2 Dataset \cite{PAMAP2} is collected from 9 participants who wore numerous sensors, including chest, wrist, and ankle sensors, to collect data. The UCIHAR Dataset \cite{UCIHAR} comprises data from 30 participants who performed 6 activities while donning a smartphone on their midsection. The UTD-MHAD Dataset \cite{UTD-MHAD} is a high-quality action dataset gathered using a Microsoft Kinect camera and a wearable IMU sensor. It includes 27 actions, and each is performed four times by 8 individuals. Finally, the DSADS Dataset \cite{DSADS} consists of data collected on 19 activities performed for 5 minutes by eight participants, 9 of which were utilized in our evaluation.

\begin{table*}[h]
    \caption{Comparisons on contrastive learning approaches with the combination of different data augmentation functions (O: Original Data, J: Jittering, P: Permutation, S: Scaling) on PAMAP2, UCIHAR, UTD-MHAD, DSADS.}
    \label{tab:exp_pamap_uci}
    \footnotesize
    \renewcommand\arraystretch{1.2}
    \begin{minipage}{.5\textwidth}
        \captionsetup{font={small}}
        \caption*{Dataset: PAMAP2}
        \vspace{-0.2cm}
        \centering
        \setlength\tabcolsep{1mm}{
        \begin{tabular}{ccccccc} \toprule 
        \multirow{2}[2]{*}{Method} & \multicolumn{6}{c} {Augmentation Strategies} \cr
        \cmidrule(lr){2-7} & \footnotesize{O, J} & \footnotesize{O, P} & \footnotesize{O, S} & \footnotesize{P, J} & \footnotesize{S, J} & \footnotesize{S, P}\\
        \midrule
        NNCLR
        & $84.83\%$ & $85.35\%$& $85.23\%$& $85.28\%$& $85.46\%$& $85.40\%$ \\
        
        SimCLR
        & $85.36\%$ & $86.11\%$& $84.65\%$& $86.36\%$& $85.91\%$& $87.28\%$ \\
        
        SimSiam
        & $50.72\%$ & $52.26\%$& $55.42\%$& $51.81\%$& $52.89\%$& $55.53\%$ \\
        
        BYOL
        & $88.97\%$ & $89.93\%$& $89.26\%$& $90.31\%$& $90.00\%$& $90.30\%$ \\
        \hline
        \hline
        \multirow{2}[2]{*}{AutoCL} & \multicolumn{6}{c} {Auto-Augmentation} \cr
        \cmidrule(lr){2-7}
        & \multicolumn{6}{c}{\textbf{90.47\%}} \cr
        \bottomrule
        \end{tabular}
        }
    \end{minipage}
    \begin{minipage}{.5\textwidth}
        \centering
        \captionsetup{font={small}}
        \caption*{Dataset: UCIHAR}
        \vspace{-0.2cm}
        \setlength\tabcolsep{1mm}{
        \begin{tabular}{ccccccc} \toprule 
        \multirow{2}[2]{*}{Method} & \multicolumn{6}{c} {Augmentation Strategies} \cr
        \cmidrule(lr){2-7} & \footnotesize{O, J} & \footnotesize{O, P} & \footnotesize{O, S} & \footnotesize{P, J} & \footnotesize{S, J} & \footnotesize{S, P}\\
        \midrule
        NNCLR
        & $90.95\%$ & $89.63\%$& $89.70\%$& $92.02\%$& $90.98\%$& $90.03\%$ \\
        
        SimCLR
        & $91.22\%$ & $94.18\%$& $84.65\%$& $86.36\%$& $85.91\%$& $87.28\%$ \\
        
        SimSiam
        & $84.44\%$ & $86.54\%$& $80.33\%$& $82.83\%$& $86.12\%$& $83.38\%$ \\
        
        BYOL
        & $91.82\%$ & $93.84\%$& $92.67\%$& $92.29\%$& $91.25\%$& $92.52\%$ \\
        \hline
        \hline
        \multirow{2}[2]{*}{AutoCL} & \multicolumn{6}{c} {Auto-Augmentation} \cr
        \cmidrule(lr){2-7}
        & \multicolumn{6}{c}{\textbf{94.69\%}} \cr
        \bottomrule
        \end{tabular}
        }
    \end{minipage} 
\end{table*}
\begin{table*}[h]
    \label{tab:exp_utd_DSADS}
    \footnotesize
    \renewcommand\arraystretch{1.2}
    \begin{minipage}{.5\textwidth}
        \captionsetup{font={small}}
        \caption*{Dataset: UTD-MHAD}
        \vspace{-0.2cm}
        \centering
        \setlength\tabcolsep{1mm}{
        \begin{tabular}{ccccccc} \toprule 
        \multirow{2}[2]{*}{Method} & \multicolumn{6}{c} {Augmentation Strategies} \cr
        \cmidrule(lr){2-7} & \footnotesize{O, J} & \footnotesize{O, P} & \footnotesize{O, S} & \footnotesize{P, J} & \footnotesize{S, J} & \footnotesize{S, P}\\
        \midrule
        NNCLR
        & $60.30\%$ & $61.71\%$& $62.34\%$& $63.76\%$& $62.54\%$& $62.99\%$ \\
        
        SimCLR
        & $55.09\%$ & $60.56\%$& $58.46\%$& $61.03\%$& $57.51\%$& $60.80\%$ \\
        
        SimSiam
        & $56.27\%$ & $54.86\%$& $56.64\%$& $56.54\%$& $60.13\%$& $57.12\%$ \\
        
        BYOL
        & $59.93\%$ & $58.71\%$& $59.13\%$& $59.13\%$& $58.28\%$& $57.99\%$ \\
        \hline
        \hline
        \multirow{2}[2]{*}{AutoCL} & \multicolumn{6}{c} {Auto-Augmentation} \cr
        \cmidrule(lr){2-7}
        & \multicolumn{6}{c}{\textbf{64.51\%}} \cr
        
        \bottomrule
        \end{tabular}
        }
    \end{minipage}
    \begin{minipage}{.5\textwidth}
        \centering
        \captionsetup{font={small}}
        \caption*{Dataset: DSADS}
        \vspace{-0.2cm}
        \setlength\tabcolsep{1mm}{
        \begin{tabular}{ccccccc} \toprule 
        \multirow{2}[2]{*}{Method} & \multicolumn{6}{c} {Augmentation Strategies} \cr
        \cmidrule(lr){2-7} & \footnotesize{O, J} & \footnotesize{O, P} & \footnotesize{O, S} & \footnotesize{P, J} & \footnotesize{S, J} & \footnotesize{S, P}\\
        \midrule
        NNCLR
        & $91.17\%$ & $91.42\%$& $92.54\%$& $91.41\%$& $92.42\%$& $91.98\%$ \\
        
        SimCLR
        & $91.37\%$ & $84.63\%$& $89.24\%$& $87.62\%$& $93.64\%$& $86.20\%$ \\
        
        SimSiam
        & $47.79\%$ & $50.68\%$& $53.03\%$& $55.79\%$& $56.82\%$& $59.29\%$ \\
        
        BYOL
        & $93.78\%$ & $94.23\%$& $95.59\%$& $94.86\%$& $95.77\%$& $94.41\%$ \\
        \hline
        \hline
        \multirow{2}[2]{*}{AutoCL} & \multicolumn{6}{c} {Auto-Augmentation} \cr
        \cmidrule(lr){2-7}
        & \multicolumn{6}{c}{\textbf{96.05\%}} \cr
        \bottomrule
        \end{tabular}
        }
    \end{minipage} 
    \vspace{-0.3cm}
\end{table*}

\subsection{Experimental Setup}\label{sec:ExpSetup}



In our experiment, we employ contrastive learning methods for pre-training the encoder, followed by the few-shot classification task as a downstream evaluation method, following the evaluation framework in \cite{SSLTimeSeries}. During the pre-training phase, we implement four state-of-the-art contrastive learning methods as baselines utilizing the same backbone for training without labels. The same batch size, optimizer, and procedure are used to facilitate comparison. In evaluation, we add the MLP prediction module to the pre-trained encoder for datasets' classification by fixing the pre-trained encoder and fine-tuning the prediction module. During the fine-tuning process, we randomly choose 20\% of the labeled data from the dataset for few-shot learning. The remaining 80\% of the data is used to evaluate the classification accuracy, thereby further measuring the performance of different baselines. The details of the pre-training and fine-tuning setup are as follows:

\subsubsection{Pre-Training}
The pre-training employs the 3-layer FCN as the encoder backbone network (Section \ref{sec:encoder}) and maintains an identical projection module (Section \ref{sec:projector}). The batch size is set at 256 (Section \ref{sec:optimization}). To ensure a consistent model selection strategy, early stopping is adopted; pre-training is halted after five epochs without loss decrease, with the model corresponding to the smallest loss selected for subsequent evaluation.

\subsubsection{Fine-Tuning}
During fine-tuning, the prediction module comprises two fully-connected layers, with batch normalization and ReLU activation in the first, and softmax activation in the second for classification. The number of neurons in these layers is 128 and the number of classes, respectively. A batch size of 128 is adopted, with the Adam optimizer set at a learning rate of $\num{1e-3}$ and weight decay of $\num{1e-4}$. To normalize fine-tuning evaluation, each method is fine-tuned 100 epochs. The final evaluation accuracy is determined by averaging the top ten accuracy results from the test set.

\subsection{Comparison with Other Methods}\label{sec:comparison}
We aim to compare AutoCL with state-of-the-art contrastive learning baselines such as NNCLR \cite{NNCLR}, SimCLR \cite{SimCLR}, SimSiam \cite{SimSiam}, and BYOL \cite{BYOL}. These contrastive learning methods are all based on a dual-stream network structure, with implementations based on the same backbone (Section \ref{sec:ImplementationDetails}) and hyperparameters being the same. Their inputs depend on the data augmentation strategy. Therefore, we choose to use a combination of time-series data augmentation methods, which are widely acknowledged in the HAR \cite{TSTCC} for better evaluation:

\begin{itemize}

\item Jittering (J): Adds random Gaussian noise to signals.

\item Scaling (S): An augmentation that multiplies input signals with values sampled from the normal distribution.

\item Permutation (P): Splits input signals into a certain number of intervals and randomly permutes them.

\item Original Data (O): Original data without data augmentation.

\end{itemize}

In our evaluation, we compare AutoCL, which does not require data augmentation operations in the pretext task, with state-of-the-art contrastive learning baselines that adopt a combination of the augmentation strategies. The results indicate that the optimal data augmentation method or combination is inconsistent across different datasets from varying scenarios, modalities, and data collection settings. AutoCL not only releases the burden of data augmentation under various applications but also achieves state-of-the-art performance on these HAR datasets, as detailed in Table \ref{tab:exp_pamap_uci}. This suggests that its design approach is effective.

\begin{figure}[htbp]
    \vspace{-0.3cm}
    \centering
    \includegraphics[scale=0.34]{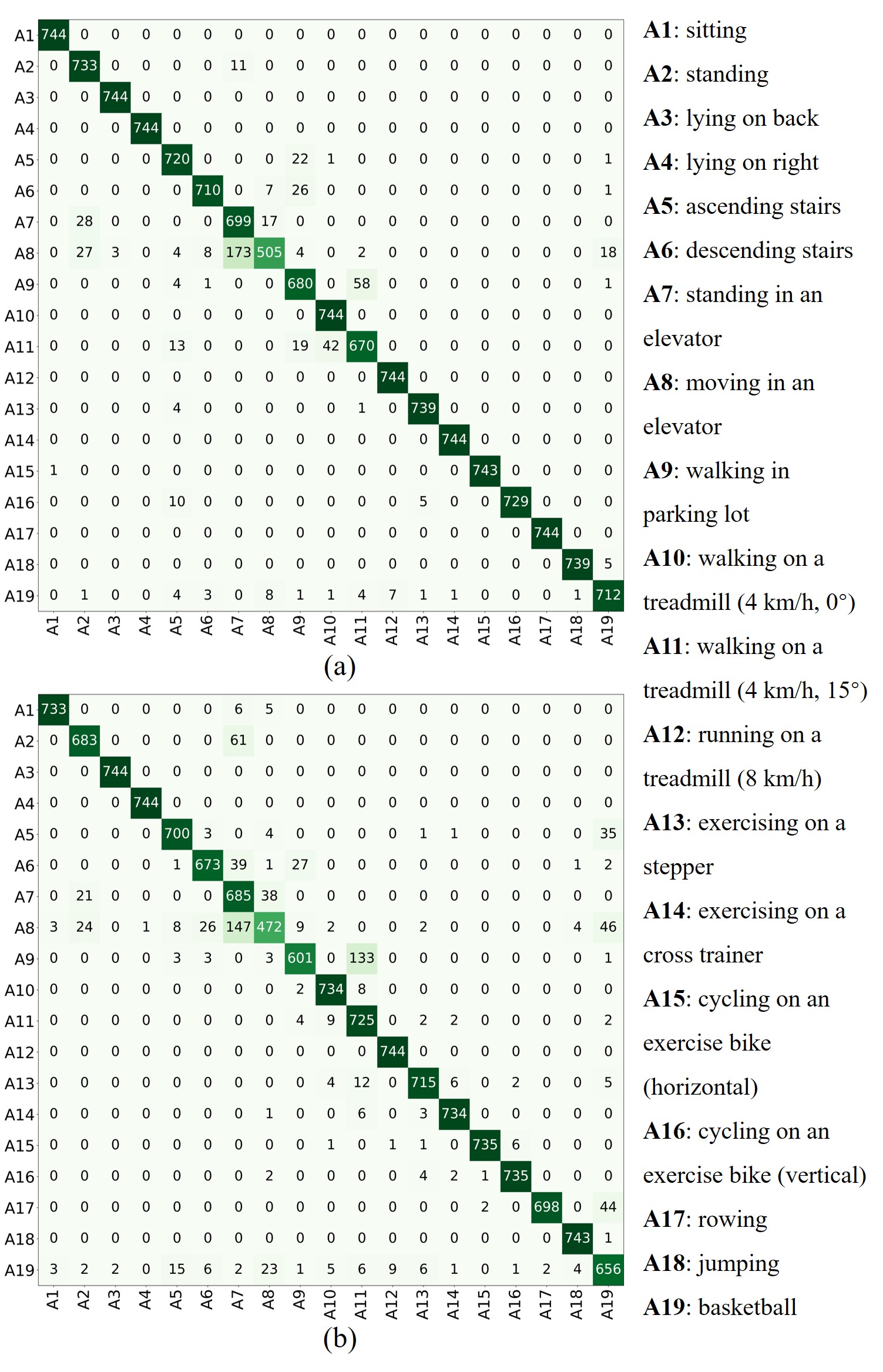}
    \vspace{-0.3cm}
    \caption{Illustrate confusion matrix: (a) AutoCL. (b) SimCLR.}
    \label{fig:confusion}
    \vspace{-0.3cm}
\end{figure}

To further analyze the effectiveness of the AutoCL mechanism, since AutoCL is mainly based on the SimCLR architecture, we use SimCLR as a baseline with the best augmentation strategy in Table \ref{tab:exp_pamap_uci} to show the confusion matrix with AutoCL on DSADS. DSADS has a more significant number of samples, which can indicate the performance between the methods more clearly. As illustrated in Figure \ref{fig:confusion}, AutoCL is better regarding the number of correct categorizations. And in some activities where the difference is not significant such as A10 and A11, A9 and A11, AutoCL shows fewer errors, which implies that AutoCL has an advantage in recognizing fine-grained features in self-supervised contrastive learning due to the design of auto-augmentation.


\begin{figure}[hbtp]
    \vspace{-0.4cm}
    \centering
    \includegraphics[scale=0.35]{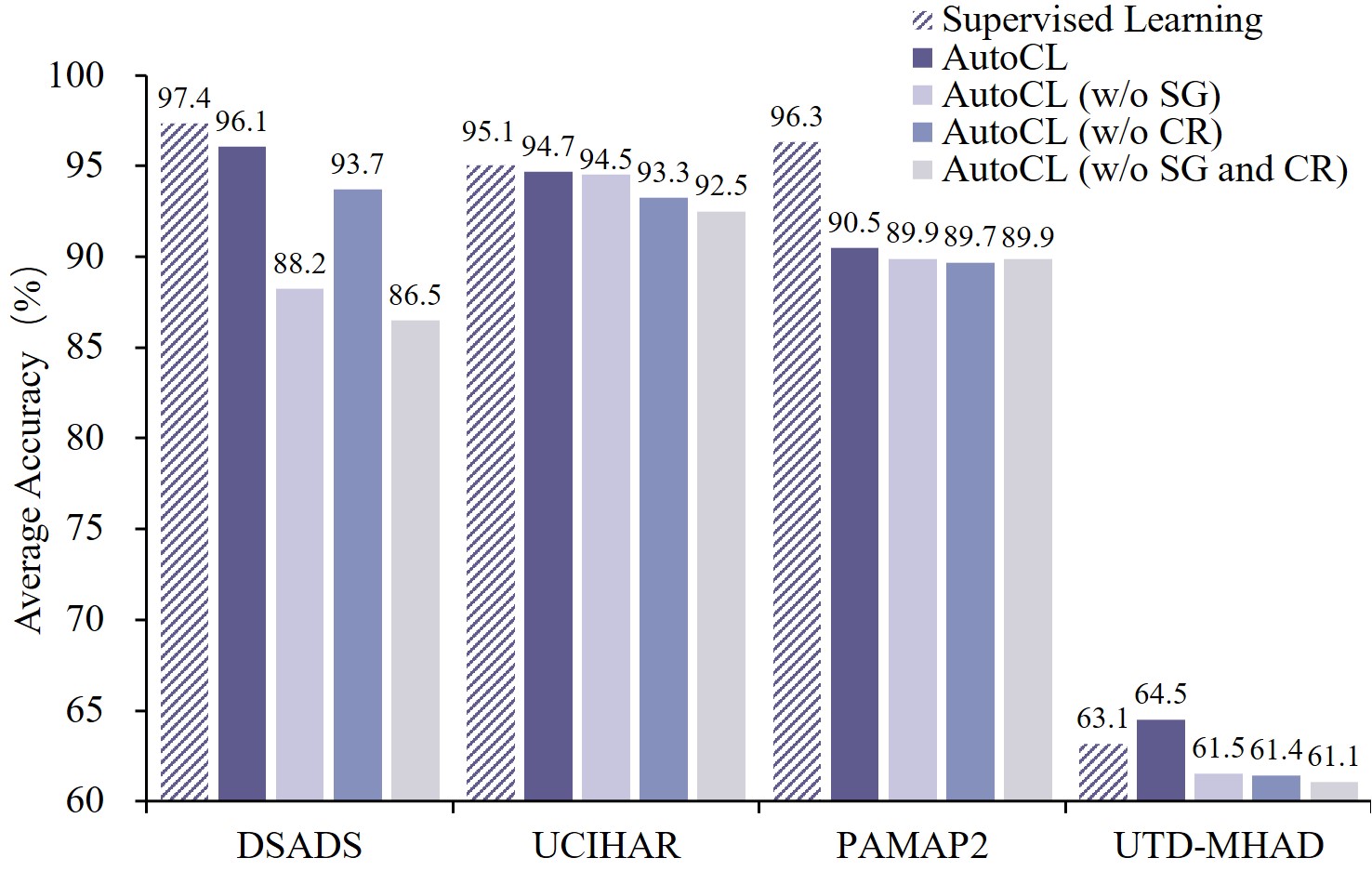}
    \vspace{-0.1cm}
    \caption{AutoCL with vs. without Stop-Gradient (SG) and/or Correlation Reduction (CR).}
    \label{fig:ablation}
    \vspace{-0.3cm}
\end{figure}

\subsection{Ablation Study}
In addition to the overall evaluation of auto-augmentation architecture, we design the ablation experiment to verify the effectiveness of the correlation reduction strategy and the stop-gradient design in AutoCL. As illustrated in Figure \ref{fig:ablation}, we compared the implementation of the AutoCL method with the alternatives that do not employ the correlation reduction (CR) strategy and the stop-gradient (SG) design. Due to the diversity among the datasets, the improvement of these strategies is not equal across datasets, especially on the UTD-MHAD dataset, where the addition of SG and CR makes the AutCL self-supervised encoder performance significantly outperform the supervised one. 

Overall AutoCL's SG and CR strategies show robustness and achieve performance gains on all datasets. The experimental results indicate that the correlation reduction strategy and the stop-gradient design in AutoCL are crucial to the architecture of automatic data augmentation, significantly enhancing the performance of our end-to-end contrastive learning approach.

\section{Discussion}
In this section, we discuss the effectiveness of the architecture design of AutoCL by conducting an architecture empirical study. Furthermore, we visually analyze the samples generated by AutoCL's augmentation to discuss the promise of data-driven data augmentation methods for low-semantic data.

\subsection{Architecture Empirical Study}\label{sec:EmpiricalStudy}
\begin{figure}[hbtp]
    \centering
    \vspace{-0.3cm}
    \includegraphics[scale=0.35]{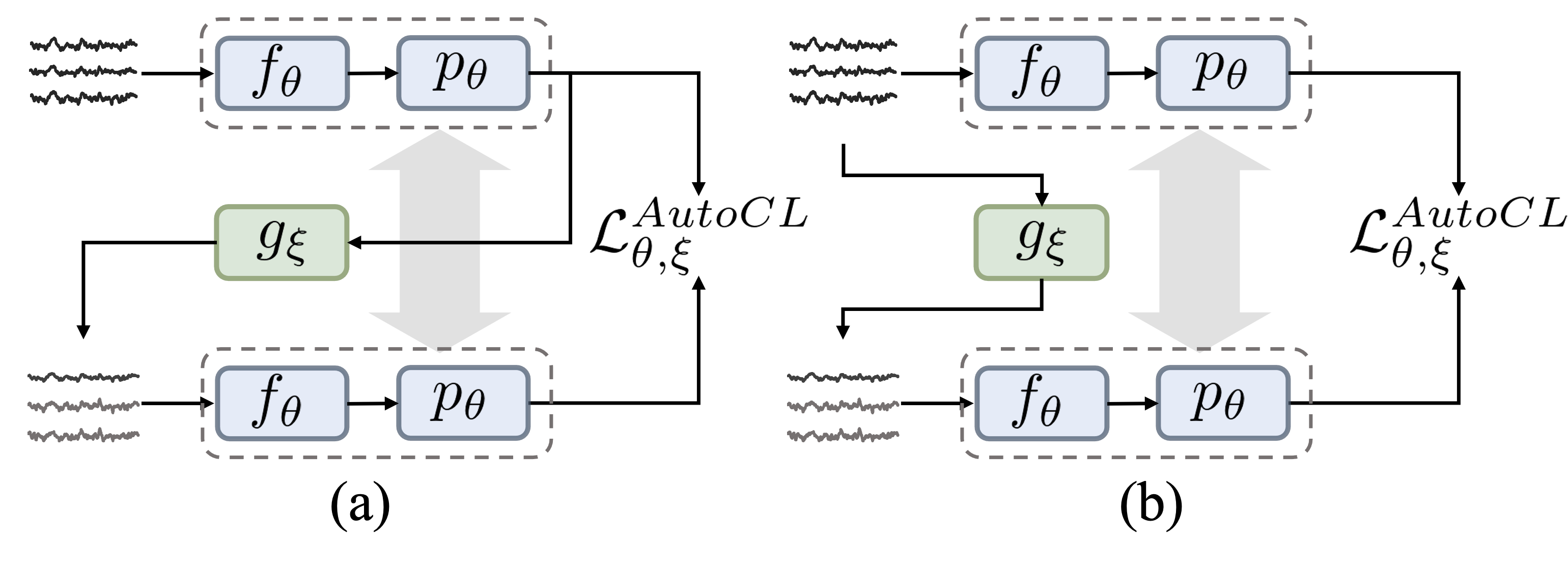}
    \vspace{-0.3cm}
    \caption{(a) AutoCL-E architecture, which is based on feature embedding. (b) AutoCL-D architecture, which is based on original data. }
    \label{fig:two_structure}
    \vspace{-0.3cm}
\end{figure}

In this section, we perform a comprehensive empirical study of the AutoCL architecture, focusing specifically on the impact of auto-augmentation derived from feature embeddings on performance enhancement. We established two auto-augmentation models: AutoCL-D, which is constructed directly from the original data (Figure \ref{fig:two_structure}(a)), and AutoCL-E, which is based on feature embedding (Figure \ref{fig:two_structure}(b)). Both models utilize the Siamese network structure and generator backbone as same as AutoCL. Notably, the input data for AutoCL-D undergoes batch normalization, while the input data for AutoCL-E is processed as detailed in Section \ref{sec:generator} due to the heterogeneity of the raw data and feature embedding. Our experimental methodology involves pre-training and few-shot classification assessments on the HAR dataset (refer to Section \ref{sec:ExpSetup}).

\begin{figure}[htbp]
    \vspace{-0.3cm}
    \centering
    \includegraphics[scale=0.08]{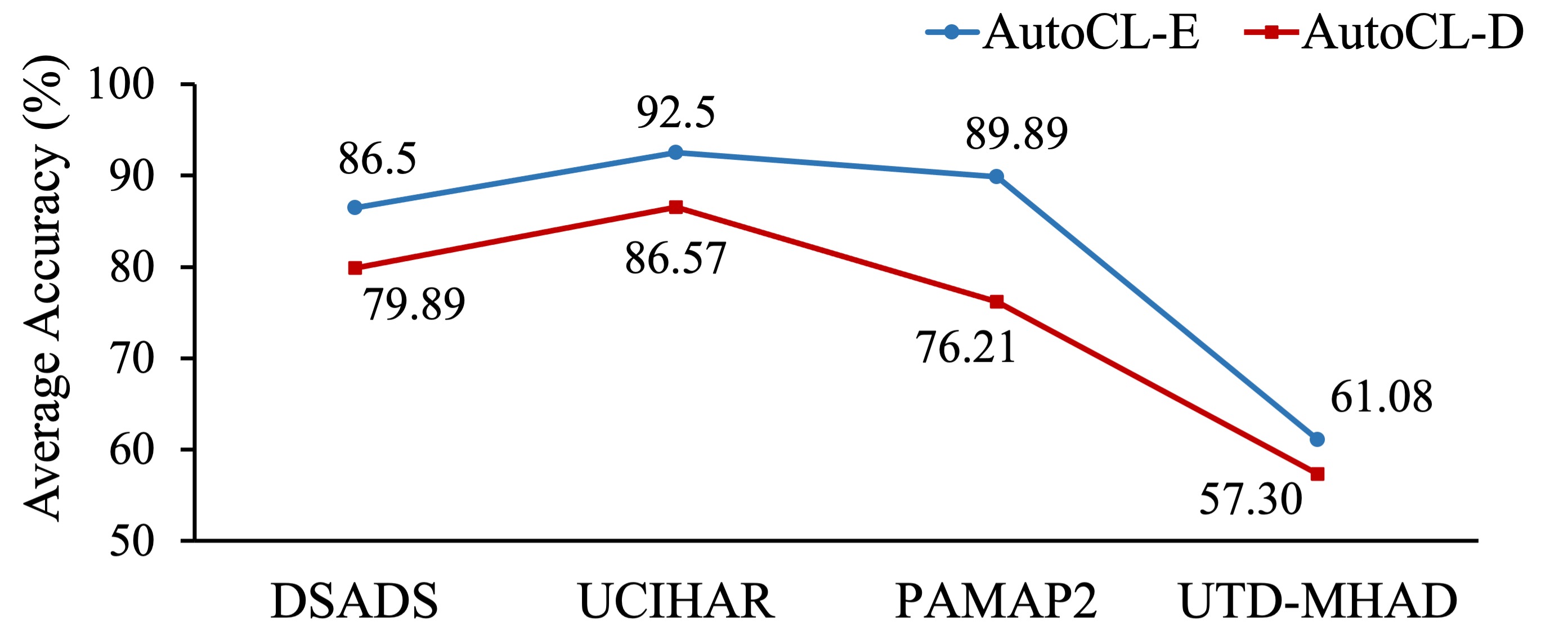}
    \vspace{-0.1cm}
    \caption{Comparison of Auto-E and Auto-D architecture.}
    \label{fig:architecture_exp}
    \vspace{-0.3cm}
\end{figure}

The results of these evaluations are presented in Figure \ref{fig:architecture_exp}. AutoCL-E is generally superior to AutoCL-D. The PAMAP2, UCIHAR, and DSADS consist of continuous activities with considerable variations. In contrast, UTD-MHAD, characterized by atomic actions following the same trajectory, exhibits less data noise. As a result, the performance difference between AutoCL-E and AutoCL-D on the UTD-MHAD dataset is the smallest.

\begin{figure}[htbp]
    \centering
    \includegraphics[scale=0.35]{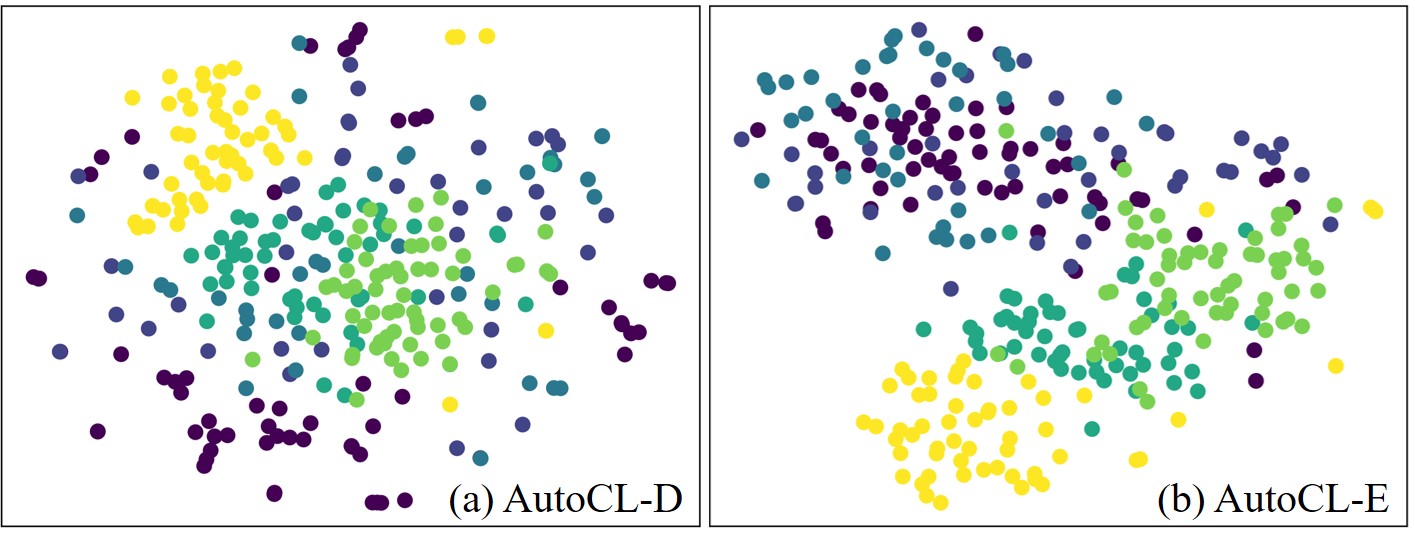}
    \vspace{-0.1cm}
    \caption{The t-SNE diagrams of the pre-trained encoders of AutoCL-D and AutoCL-E.}
    \label{fig:tSNE}
    \vspace{-0.3cm}
\end{figure}

Moreover, we display the t-distributed stochastic neighborhood embedding (t-SNE) diagrams of the pre-trained encoders of the two architectures on the UCI dataset in Figure \ref{fig:tSNE}. The UCI dataset has minimal categories for easy understanding. These diagrams illustrate that auto-augmentation via feature embeddings enhances the representation capacity of the contrastive learning encoder. For instance, Figure \ref{fig:tSNE}(a) reveals some difficulties in distinguishing features except the yellow class, whereas \ref{fig:tSNE}(b) demonstrates that the features generated by AutoCL-E are more discernible.

The architecture empirical study results demonstrate that AutoCL-E's embedding-feature-based auto-augmentation architecture can significantly improve performance. For HAR sensor data, auto-augmentation based on latent space representation can facilitate the backbone to learn effective representations and avoid sensor noise and redundant information interference. The empirical study validates the rationality and effectiveness of the design motivation for the AutoCL architecture in Section \ref{sec:Motivation}.

\subsection{Auto-Augmentation Visualize}
\begin{figure}[htbp]
   \centering
   \includegraphics[scale=0.35]{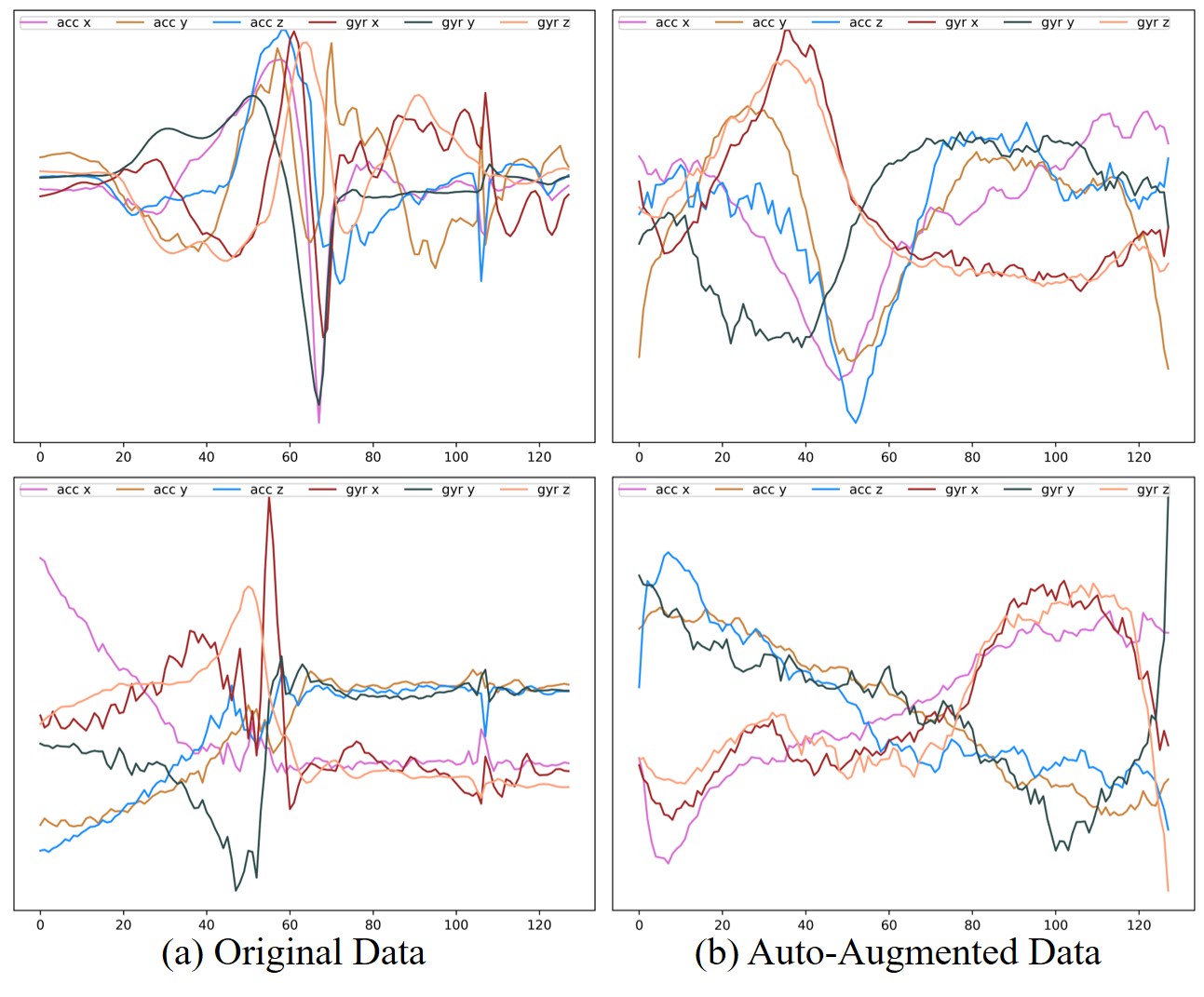}
   \vspace{-0.3cm}
   \caption{Visualization of original and auto-augmented data using the 6-axis IMU sample from UTD-MHAD dataset.}
   \label{fig:visualize}
   \vspace{-0.3cm}
\end{figure}

To delve deeper into the details and principles of the AutoCL design, we have visualized the samples generated by the generator compared to the original samples, as shown in Figure \ref{fig:visualize}. We visualize the original and auto-augmented 6-axis IMU data from UTD-MHAD. It can be seen that AutoCL implements augmentation for the samples, consisting mainly of operations such as scaling, cropping, reversing, etc. This data augmentation is data-driven. The model will implement data-driven augmentation for the raw data, e.g., the first line mainly performs scaling, and the second line mainly performs reversing, but is not limited to this. For low semantic sensor data, the space of data-driven augmentation operations should be far more than the human-specified strategy, with the capability to achieve better performance.

%

\section{Conclution}
Human activity recognition has become ubiquitous in recent years due to mobile and wearable computing advancements. In the context of prevalent low-semantic sensor signals, pursuing high performance and reduced manual annotation burden has spurred research and development in contrastive learning within this domain. However, data augmentation in the pretext tasks of contrastive learning remains, to some extent, manual and non-data-driven, lacking generalizability and flexibility. Therefore, we propose AutoCL, an end-to-end auto-augmentation contrastive learning method that implements high-performance contrastive learning through auto-augmentation architecture, correlation reduction strategy, and stop-gradient design. Furthermore, for low-semantic data, a data-driven automated augmentation approach holds greater space for data augmentation operations than the human-specified strategy and has more potential to ensure semantic consistency with flexibility between augmented samples than manually implemented.

\section{Acknowledgments}

This work is supported by the National Key Research \& Development Plan of China No. 2021YFC2501202, National Natural Science Foundation of China (No. 62202455, No. 61972383, No. 62101530), Youth Innovation Promotion Association CAS (No. 2021101), Beijing Municipal Science \& Technology Commission (No. Z221100002722009).

\bibliographystyle{IEEEtran}
\bibliography{bibfile}

@inproceedings{PAMAP2,
  title={Introducing a new benchmarked dataset for activity monitoring},
  author={Reiss, Attila and Stricker, Didier},
  booktitle={2012 16th international symposium on wearable computers},
  pages={108--109},
  year={2012},
  organization={IEEE}
}

@inproceedings{UCIHAR,
  title={Human activity recognition on smartphones using a multiclass hardware-friendly support vector machine},
  author={Anguita, Davide and Ghio, Alessandro and Oneto, Luca and Parra, Xavier and Reyes-Ortiz, Jorge L},
  booktitle={Ambient Assisted Living and Home Care: 4th International Workshop, IWAAL 2012, Vitoria-Gasteiz, Spain, December 3-5, 2012. Proceedings 4},
  pages={216--223},
  year={2012},
  organization={Springer}
}

@inproceedings{CL-HAR,
  title={What makes good contrastive learning on small-scale wearable-based tasks?},
  author={Qian, Hangwei and Tian, Tian and Miao, Chunyan},
  booktitle={Proceedings of the 28th ACM SIGKDD Conference on Knowledge Discovery and Data Mining},
  pages={3761--3771},
  year={2022}
}

@inproceedings{SimCLR,
  title={A simple framework for contrastive learning of visual representations},
  author={Chen, Ting and Kornblith, Simon and Norouzi, Mohammad and Hinton, Geoffrey},
  booktitle={International conference on machine learning},
  pages={1597--1607},
  year={2020},
  organization={PMLR}
}

@article{COCOA,
  title={Cocoa: Cross modality contrastive learning for sensor data},
  author={Deldari, Shohreh and Xue, Hao and Saeed, Aaqib and Smith, Daniel V and Salim, Flora D},
  journal={Proceedings of the ACM on Interactive, Mobile, Wearable and Ubiquitous Technologies},
  volume={6},
  number={3},
  pages={1--28},
  year={2022},
  publisher={ACM New York, NY, USA}
}

@inproceedings{CLvsAE,
  title={Contrastive self-supervised learning for sensor-based human activity recognition},
  author={Khaertdinov, Bulat and Ghaleb, Esam and Asteriadis, Stylianos},
  booktitle={2021 IEEE International Joint Conference on Biometrics (IJCB)},
  pages={1--8},
  year={2021},
  organization={IEEE}
}

@inproceedings{NNCLR,
  title={With a little help from my friends: Nearest-neighbor contrastive learning of visual representations},
  author={Dwibedi, Debidatta and Aytar, Yusuf and Tompson, Jonathan and Sermanet, Pierre and Zisserman, Andrew},
  booktitle={Proceedings of the IEEE/CVF International Conference on Computer Vision},
  pages={9588--9597},
  year={2021}
}

@inproceedings{SimSiam,
  title={Exploring simple siamese representation learning},
  author={Chen, Xinlei and He, Kaiming},
  booktitle={Proceedings of the IEEE/CVF conference on computer vision and pattern recognition},
  pages={15750--15758},
  year={2021}
}

@article{BYOL,
  title={Bootstrap your own latent-a new approach to self-supervised learning},
  author={Grill, Jean-Bastien and Strub, Florian and Altch{\'e}, Florent and Tallec, Corentin and Richemond, Pierre and Buchatskaya, Elena and Doersch, Carl and Avila Pires, Bernardo and Guo, Zhaohan and Gheshlaghi Azar, Mohammad and others},
  journal={Advances in neural information processing systems},
  volume={33},
  pages={21271--21284},
  year={2020}
}

@article{TSTCC,
  title={Time-series representation learning via temporal and contextual contrasting},
  author={Eldele, Emadeldeen and Ragab, Mohamed and Chen, Zhenghua and Wu, Min and Kwoh, Chee Keong and Li, Xiaoli and Guan, Cuntai},
  journal={arXiv preprint arXiv:2106.14112},
  year={2021}
}

@article{deeplearningHAR0,
  title={Deep learning for sensor-based activity recognition: A survey},
  author={Wang, Jindong and Chen, Yiqiang and Hao, Shuji and Peng, Xiaohui and Hu, Lisha},
  journal={Pattern recognition letters},
  volume={119},
  pages={3--11},
  year={2019},
  publisher={Elsevier}
}

@inproceedings{deeplearningHAR1,
  title={An Enhanced Human Activity Recognition Algorithm with Positional Attention},
  author={Xu, Chenyang and Shen, Jianfei and Fan, Feiyi and Qiu, Tian and Mao, Zhihong},
  booktitle={Asian Conference on Machine Learning},
  pages={1181--1196},
  year={2023},
  organization={PMLR}
}

@inproceedings{deeplearningHAR2,
  title={A Shallow Convolution Network Based Contextual Attention for Human Activity Recognition},
  author={Xu, Chenyang and Mao, Zhihong and Fan, Feiyi and Qiu, Tian and Shen, Jianfei and Gu, Yang},
  booktitle={International Conference on Mobile and Ubiquitous Systems: Computing, Networking, and Services},
  pages={155--171},
  year={2022},
  organization={Springer}
}

@inproceedings{semilearning0,
  title={LabelForest: Non-parametric semi-supervised learning for activity recognition},
  author={Ma, Yuchao and Ghasemzadeh, Hassan},
  booktitle={Proceedings of the AAAI Conference on Artificial Intelligence},
  volume={33},
  number={01},
  pages={4520--4527},
  year={2019}
}

@article{semilearning1,
  title={Study on human activity recognition using semi-supervised active transfer learning},
  author={Oh, Seungmin and Ashiquzzaman, Akm and Lee, Dongsu and Kim, Yeonggwang and Kim, Jinsul},
  journal={Sensors},
  volume={21},
  number={8},
  pages={2760},
  year={2021},
  publisher={MDPI}
}

@article{weaklearning,
  title={Weakly-supervised sensor-based activity segmentation and recognition via learning from distributions},
  author={Qian, Hangwei and Pan, Sinno Jialin and Miao, Chunyan},
  journal={Artificial Intelligence},
  volume={292},
  pages={103429},
  year={2021},
  publisher={Elsevier}
}

@inproceedings{weaklearning1,
  title={Siamese networks for weakly supervised human activity recognition},
  author={Sheng, Taoran and Huber, Manfred},
  booktitle={2019 IEEE International Conference on Systems, Man and Cybernetics (SMC)},
  pages={4069--4075},
  year={2019},
  organization={IEEE}
}

@inproceedings{translearning0,
  title={Latent independent excitation for generalizable sensor-based cross-person activity recognition},
  author={Qian, Hangwei and Pan, Sinno Jialin and Miao, Chunyan},
  booktitle={Proceedings of the AAAI Conference on Artificial Intelligence},
  volume={35},
  number={13},
  pages={11921--11929},
  year={2021}
}

@inproceedings{ssl0,
  title={Masked reconstruction based self-supervision for human activity recognition},
  author={Haresamudram, Harish and Beedu, Apoorva and Agrawal, Varun and Grady, Patrick L and Essa, Irfan and Hoffman, Judy and Pl{\"o}tz, Thomas},
  booktitle={Proceedings of the 2020 ACM International Symposium on Wearable Computers},
  pages={45--49},
  year={2020}
}

@article{augmentationmethods,
  title={Exploring contrastive learning in human activity recognition for healthcare},
  author={Tang, Chi Ian and Perez-Pozuelo, Ignacio and Spathis, Dimitris and Mascolo, Cecilia},
  journal={arXiv preprint arXiv:2011.11542},
  year={2020}
}

@inproceedings{stfcl,
  title={Contrastive self-supervised representation learning for sensing signals from the time-frequency perspective},
  author={Liu, Dongxin and Wang, Tianshi and Liu, Shengzhong and Wang, Ruijie and Yao, Shuochao and Abdelzaher, Tarek},
  booktitle={2021 International Conference on Computer Communications and Networks (ICCCN)},
  pages={1--10},
  year={2021},
  organization={IEEE}
}

@inproceedings{yao2019stfnets,
  title={Stfnets: Learning sensing signals from the time-frequency perspective with short-time fourier neural networks},
  author={Yao, Shuochao and Piao, Ailing and Jiang, Wenjun and Zhao, Yiran and Shao, Huajie and Liu, Shengzhong and Liu, Dongxin and Li, Jinyang and Wang, Tianshi and Hu, Shaohan and others},
  booktitle={The World Wide Web Conference},
  pages={2192--2202},
  year={2019}
}

@article{augmentationfree,
  title={Augmentation-Free Graph Contrastive Learning of Invariant-Discriminative Representations},
  author={Li, Haifeng and Cao, Jun and Zhu, Jiawei and Luo, Qinyao and He, Silu and Wang, Xuying},
  journal={IEEE Transactions on Neural Networks and Learning Systems},
  year={2023},
  publisher={IEEE}
}

@article{studyCL0,
  title={What makes for good views for contrastive learning?},
  author={Tian, Yonglong and Sun, Chen and Poole, Ben and Krishnan, Dilip and Schmid, Cordelia and Isola, Phillip},
  journal={Advances in neural information processing systems},
  volume={33},
  pages={6827--6839},
  year={2020}
}

@article{studyCL1,
  title={Demystifying contrastive self-supervised learning: Invariances, augmentations and dataset biases},
  author={Purushwalkam, Senthil and Gupta, Abhinav},
  journal={Advances in Neural Information Processing Systems},
  volume={33},
  pages={3407--3418},
  year={2020}
}

@article{studyCL2,
  title={When does contrastive learning preserve adversarial robustness from pretraining to finetuning?},
  author={Fan, Lijie and Liu, Sijia and Chen, Pin-Yu and Zhang, Gaoyuan and Gan, Chuang},
  journal={Advances in neural information processing systems},
  volume={34},
  pages={21480--21492},
  year={2021}
}

@article{reviewCL,
  title={Contrastive representation learning: A framework and review},
  author={Le-Khac, Phuc H and Healy, Graham and Smeaton, Alan F},
  journal={Ieee Access},
  volume={8},
  pages={193907--193934},
  year={2020},
  publisher={IEEE}
}

@inproceedings{clusterCL0,
  title={With a little help from my friends: Nearest-neighbor contrastive learning of visual representations},
  author={Dwibedi, Debidatta and Aytar, Yusuf and Tompson, Jonathan and Sermanet, Pierre and Zisserman, Andrew},
  booktitle={Proceedings of the IEEE/CVF International Conference on Computer Vision},
  pages={9588--9597},
  year={2021}
}

@inproceedings{UTD-MHAD,
  title={UTD-MHAD: A multimodal dataset for human action recognition utilizing a depth camera and a wearable inertial sensor},
  author={Chen, Chen and Jafari, Roozbeh and Kehtarnavaz, Nasser},
  booktitle={2015 IEEE International conference on image processing (ICIP)},
  pages={168--172},
  year={2015},
  organization={IEEE}
}

@misc{DSADS,
  author       = {Barshan,Billur and Altun,Kerem},
  title        = {{Daily and Sports Activities}},
  year         = {2013},
  howpublished = {UCI Machine Learning Repository},
  note         = {{DOI}: https://doi.org/10.24432/C5C59F}
}

@inproceedings{SensoryGANs,
  title={SensoryGANs: An effective generative adversarial framework for sensor-based human activity recognition},
  author={Wang, Jiwei and Chen, Yiqiang and Gu, Yang and Xiao, Yunlong and Pan, Haonan},
  booktitle={2018 International Joint Conference on Neural Networks (IJCNN)},
  pages={1--8},
  year={2018},
  organization={IEEE}
}

@article{SensoryGANs1,
  title={A wearable-HAR oriented sensory data generation method based on spatio-temporal reinforced conditional GANs},
  author={Wang, Jiwei and Chen, Yiqiang and Gu, Yang},
  journal={Neurocomputing},
  volume={493},
  pages={548--567},
  year={2022},
  publisher={Elsevier}
}

@inproceedings{SE,
  title={Squeeze-and-excitation networks},
  author={Hu, Jie and Shen, Li and Sun, Gang},
  booktitle={Proceedings of the IEEE conference on computer vision and pattern recognition},
  pages={7132--7141},
  year={2018}
}

@inproceedings{SSLTimeSeries,
  title={Self-supervised pre-training for time series classification},
  author={Shi, Pengxiang and Ye, Wenwen and Qin, Zheng},
  booktitle={2021 International Joint Conference on Neural Networks (IJCNN)},
  pages={1--8},
  year={2021},
  organization={IEEE}
}

@inproceedings{InstDisc,
  title={Unsupervised feature learning via non-parametric instance discrimination},
  author={Wu, Zhirong and Xiong, Yuanjun and Yu, Stella X and Lin, Dahua},
  booktitle={Proceedings of the IEEE conference on computer vision and pattern recognition},
  pages={3733--3742},
  year={2018}
}

@inproceedings{moco,
  title={Momentum contrast for unsupervised visual representation learning},
  author={He, Kaiming and Fan, Haoqi and Wu, Yuxin and Xie, Saining and Girshick, Ross},
  booktitle={Proceedings of the IEEE/CVF conference on computer vision and pattern recognition},
  pages={9729--9738},
  year={2020}
}

@inproceedings{FCN,
  title={Fully convolutional networks for semantic segmentation},
  author={Long, Jonathan and Shelhamer, Evan and Darrell, Trevor},
  booktitle={Proceedings of the IEEE conference on computer vision and pattern recognition},
  pages={3431--3440},
  year={2015}
}

\end{document}